\documentclass[acmtog]{acmart}
\usepackage{booktabs} 
\usepackage{multirow} 
\usepackage{algorithm}
\usepackage{algorithmic}
\usepackage{amsmath}
\usepackage{makecell} 
\usepackage{arydshln} 

\usepackage{multirow}
\usepackage{makecell}
\usepackage{booktabs}

\copyrightyear{2026}
\acmYear{2026}
\setcopyright{cc}
\setcctype{by}
\acmConference[SA Conference Papers '26]{SIGGRAPH Asia 2026 Conference Papers}{December 01--04, 2026}{Kuala Lumpur, Malaysia}
\acmBooktitle{SIGGRAPH Asia 2026 Conference Papers (SA Conference Papers '26), December 01--04, 2026, Kuala Lumpur, Malaysia}
\acmDOI{10.1145/3829340.3842163}
\acmISBN{979-8-4007-2842-6/2026/12}

\begin{document}

\title{\textbf{\textit{ShotVerse}}: Advancing Cinematic Camera Control for Text-Driven Multi-Shot Video Creation}


\author{Songlin Yang}
\authornote{These authors contributed equally to this work.}
\authornote{Project leader.}
\affiliation{\institution{MMLab@HKUST, The Hong Kong University of Science and Technology}\country{Hong Kong}}\email{syangds@connect.ust.hk}

\author{Zhe Wang}
\authornotemark[1]
\affiliation{\institution{MMLab@HKUST, The Hong Kong University of Science and Technology}\country{Hong Kong}}

\author{Xuyi Yang}
\authornotemark[1]
\affiliation{\institution{MMLab@HKUST, The Hong Kong University of Science and Technology}\country{Hong Kong}}

\author{Songchun Zhang}
\affiliation{\institution{MMLab@HKUST, The Hong Kong University of Science and Technology}\country{Hong Kong}}

\author{Xianghao Kong}
\affiliation{\institution{MMLab@HKUST, The Hong Kong University of Science and Technology}\country{Hong Kong}}

\author{Taiyi Wu}
\affiliation{\institution{Tencent Video}\country{China}}

\author{Xiaotong Zhao}
\affiliation{\institution{Tencent Video}\country{China}}

\author{Ran Zhang}
\affiliation{\institution{Tencent Video}\country{China}}

\author{Alan Zhao}
\affiliation{\institution{Tencent Video}\country{China}}

\author{Anyi Rao}
\authornote{Corresponding author.}
\affiliation{\institution{MMLab@HKUST, The Hong Kong University of Science and Technology}\country{Hong Kong}}\email{anyirao@ust.hk}


\begin{abstract}

Text-driven video generation has democratized film creation, but camera control in cinematic multi-shot scenarios remains a significant block. Implicit textual prompts lack precision, while explicit trajectory conditioning imposes prohibitive manual overhead and often triggers execution failures in current models. To overcome this bottleneck, we propose a data-centric paradigm shift, positing that aligned \textit{(Caption, Trajectory, Video)} triplets form an inherent joint distribution that can connect automated plotting and precise execution. 
Guided by this insight, we present \textbf{\textit{ShotVerse}}, a ``Plan-then-Control'' framework that decouples generation into two collaborative agents: a VLM (Vision-Language Model)-based Planner that leverages spatial priors to obtain cinematic, globally aligned trajectories from text, and a Controller that renders these trajectories into multi-shot video content via a camera adapter. Central to our approach is the construction of a data foundation: we design an automated multi-shot camera calibration pipeline aligns disjoint single-shot trajectories into a unified global coordinate system. This facilitates the curation of \textbf{\textit{ShotVerse-Bench}}, a high-fidelity cinematic dataset with a three-track evaluation protocol that serves as the bedrock for our framework. Extensive experiments demonstrate that \textbf{\textit{ShotVerse}} effectively bridges the gap between unreliable textual control and labor-intensive manual plotting, achieving superior cinematic aesthetics and generating multi-shot videos that are both camera-accurate and cross-shot consistent. 
  
\end{abstract}



\begin{teaserfigure}
\centering
  \includegraphics[width=\textwidth]{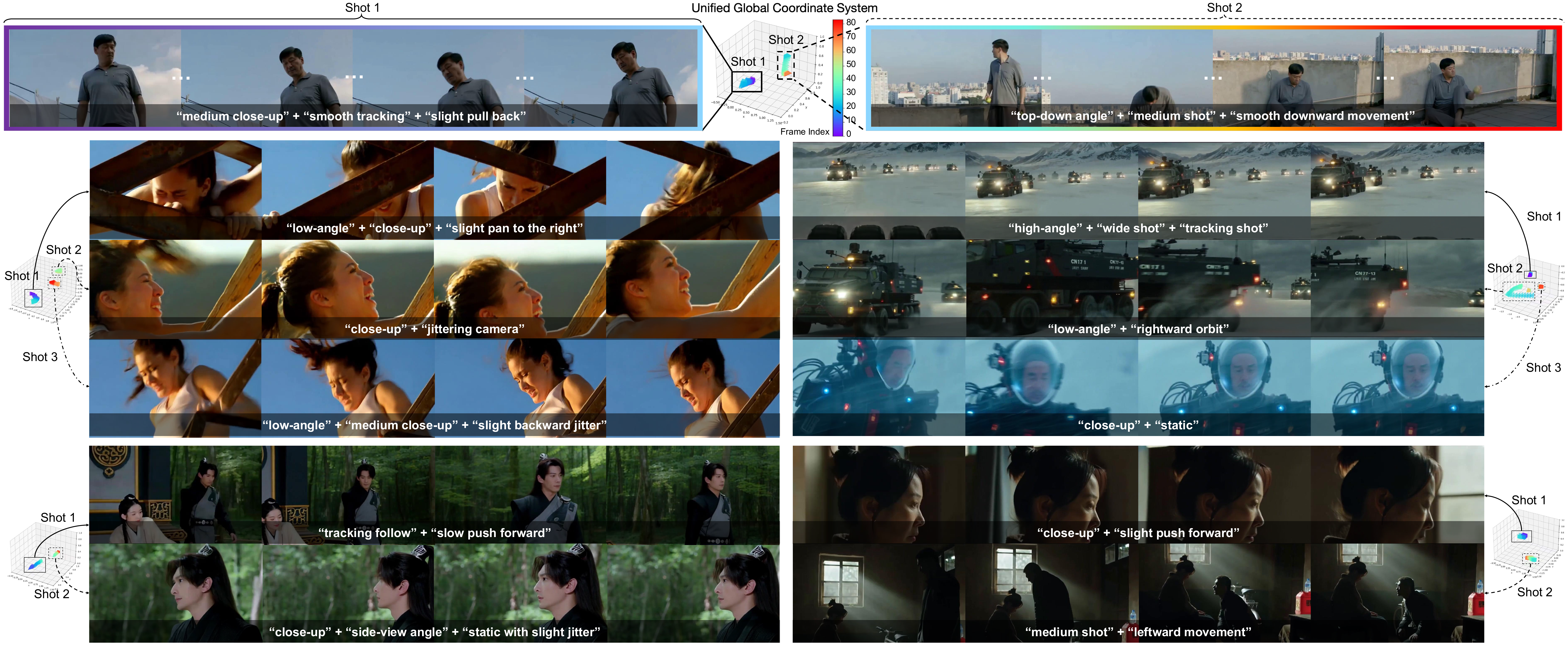}
  \vspace{-0.7cm}
  \caption{\textbf{Cinematic, Camera-Controlled, Multi-Shot Video Creation via our \textit{ShotVerse} Framework.} Examples demonstrate high-fidelity and great camera-controlled generation across diverse genres. The inset 3D plots visualize the plotted explicit trajectories.}
  \label{fig:teaser}
\end{teaserfigure}



\maketitle

\section{Introduction}

Text-driven video generation models~\cite{wan2025,hunyuanvideo,evalverse} have democratized film creation, empowering users to act as \textbf{directors} who synthesize cinematic clips from natural language. However, as the field progresses toward multi-shot video generation, while users can now easily dictate \textit{``what to see''}, the role of the \textbf{cinematographer}---executing the precise \textit{``how to shoot''} via cinematic camera control---remains a significant bottleneck~\cite{li2024can}. Existing methods either struggle to accurately follow textual camera conditions (\textit{e.g.}, ``pan left'', ``zoom in'')~\cite{holocine,multishotmaster,storymem} or fail to ensure that cameras in a multi-shot setting share a unified coordinate system~\cite{recammaster,he2024cameractrl,he2025cameractrl} and that their combination conforms to cinematic patterns~\cite{shotdirector,he2025cut2next}. A straightforward solution to these issues is to guide the video model with explicit, unified, and cinematic camera trajectories.

While this solution effectively grounds the camera motion, it introduces a new bottleneck, as plotting and executing a cinematic multi-shot trajectory is non-trivial. First, manually plotting~\cite{xing2025motioncanvas} cinematic trajectories imposes a prohibitive design burden. Users must meticulously synchronize camera poses in a unified global coordinate system with narrative flow, a labor-intensive process that demands strong spatial reasoning and intrinsic aesthetic guidance. Second, a critical execution gap remains: the state-of-the-art camera-controlled video models~\cite{recammaster} tend to treat such complex cinematic trajectories as out-of-distribution conditions, leading to failure generation (see results in Fig.~\ref{fig:compare}).

To tackle this bottleneck, we argue that this task demands a paradigm shift, rethinking from a data-centric perspective. The aligned triplets of \textit{(Caption, Trajectory, Video)} naturally form a joint distribution, and viewing the task through this lens offers two critical advantages: (i) \textit{Automating Cinematic Plotting}: By modeling the conditional probability \textit{P}(\textit{Trajectory} | \textit{Caption}), we can leverage the spatial priors of pre-trained vision-language models~\cite{cheng2024spatialrgpt,Qwen3-VL,deng2025internspatial} (VLM) to directly infer camera trajectories from text, bypassing the burden of manual plotting. (ii) \textit{Enabling Aligned-Yet-Decoupled Training}: The \textit{(Caption, Trajectory, Video)} triplets are inherently aligned, which allows decoupled optimization---optimizing \textit{P}(\textit{Trajectory} | \textit{Caption}) for plotting and \textit{P}(\textit{Video} | \textit{Caption}, \textit{Trajectory}) for generation, independently. This strategy avoids the instability of joint training while ensuring that the plotted trajectories remain compatible with the generator's execution domain.

Based on these insights, we propose \textit{\textbf{ShotVerse}}, a ``Plan-then-Control'' framework, which receives user prompts with camera description, maps them to explicit trajectories via a Planner, and utilizes a Controller to generate multi-shot videos based on these trajectories and the prompts. Specifically, \textbf{\textit{ShotVerse}} consists of two agents aligned by the shared data distribution: (i) \textit{Planner}: A VLM fine-tuned to map high-level textual descriptions into explicit camera trajectories. It leverages the learned joint distribution to predict explicit camera trajectories that respect cinematic patterns. (ii) \textit{Controller}: Built upon a holistic multi-shot video backbone~\cite{holocine}, it receives the explicit trajectories and utilizes a lightweight camera adapter to render high-fidelity cinematic content. We specifically adopt the holistic paradigm to bypass the complex inductive biases required for multi-shot consistency in auto-regressive methods~\cite{storymem,fu2026plenopticvideogeneration}, allowing us to focus purely on camera control mechanism.

Crucially, the validity of our data-centric hypothesis rests on the quality of the data itself. To address the scarcity of aligned multi-shot data, we collect raw footage from high-production videos and propose an automated camera calibration pipeline that aligns disjoint single-shot trajectories into a unified global coordinate system. Paired with hierarchical captions, this constitutes \textbf{\textit{ShotVerse-Bench}}, which serves as both a high-fidelity training foundation and a rigorous evaluation benchmark. By introducing a three-track evaluation protocol, this benchmark enables the comprehensive measurement of cinematic planning, execution fidelity, and multi-shot consistency. As shown in Fig.~\ref{fig:teaser}, extensive experiments demonstrate that our framework leverages the Planner for superior cinematic plotting and enables the Controller to generate multi-shot videos that are both camera-accurate and cross-shot consistent.

\noindent
\textbf{\textit{Distinctions and Contributions.}} (i) In contrast to full-stack filmmaking frameworks~\cite{xiao2025captain,wu2025automatedmoviegenerationmultiagent} based on multi-agent systems~\cite{dorri2018multi,liang2025univa}, \textbf{\textit{ShotVerse}} only focuses on the core component of explicit cinematic camera control. Our method bridges high-level narrative intent with precise geometry via two specialized agents: a Planner that plots script-driven, unified 3D trajectories, and a Controller that injects these professional-grade camera patterns into multi-shot foundation models. (ii) Unlike generic camera trajectory generation~\cite{gendop} and control~\cite{recammaster,he2024cameractrl}, we emphasize the automated synthesis and execution of explicit cinematic trajectories, enabling models to manifest nuanced cinematography beyond basic camera motion templates. (iii) We are the first to address explicit cinematic camera control specifically for holistic multi-shot generation models, ensuring cross-shot coherence within a unified global coordinate system. (iv) We provide an automated multi-shot camera calibration pipeline to facilitate scalable data annotation and empower future research in multi-shot video synthesis.
\vspace{-0.1cm}

\begin{figure*}[th]
    \centering
    \includegraphics[width=\linewidth]{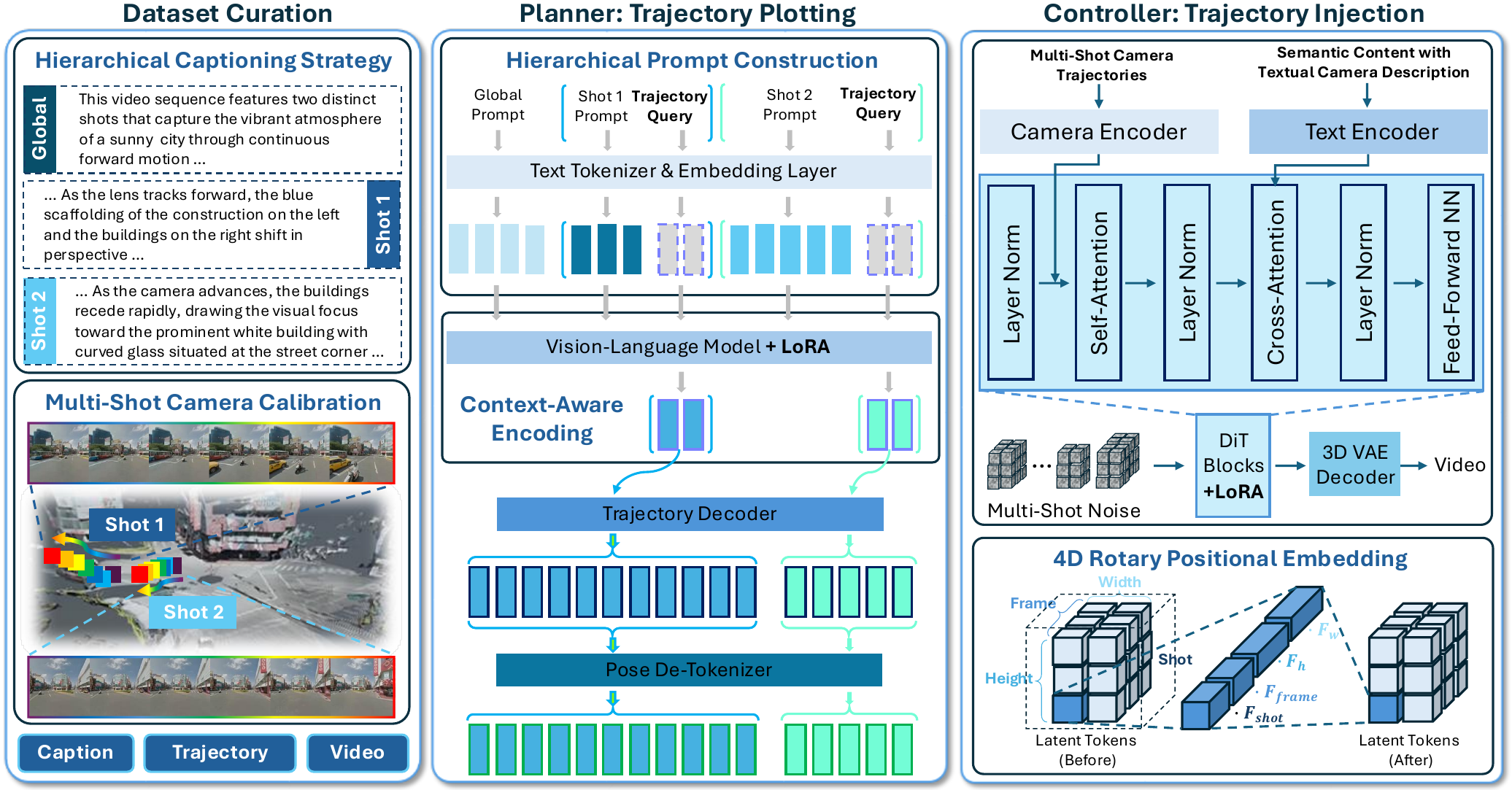}
    \vspace{-0.7cm}
    \caption{\textbf{Method Overview.} (i) \textit{Dataset Curation}. We construct the \textbf{\textit{ShotVerse-Bench}} by aligning multi-shot trajectories into a unified global coordinate system via camera calibration, paired with hierarchical global and per-shot captions. (ii) \textit{Trajectory Plotting}: The Planner utilizes a VLM to process the hierarchical prompt interleaved with learnable trajectory query tokens. These inputs are encoded into context-aware embeddings and transformed into explicit camera poses via a Trajectory Decoder and a Pose De-Tokenizer. (iii) \textit{Trajectory Injection}: The Controller synthesizes high-fidelity videos using a holistic DiT backbone. It precisely follows the trajectories via a Camera Adapter and a 4D Rotary Positional Embedding strategy.}
    \label{fig:framework}
    \vspace{-0.2cm}
\end{figure*}

\section{Related Work}

\noindent\textit{\textbf{Multi-Shot Video Creation.}} Cascaded approaches~\cite{zhou2024storydiffusion,xiao2025captain,zhang2025shouldershot,wang2025cinemaster,cinelog,zheng2025videogenofthoughtstepbystepgeneratingmultishot} struggle to maintain consistency in the temporal gaps. To address this, memory-based methods~\cite{storymem,he2025cut2next} formulate generation as iterative shot synthesis conditioned on explicit memory, yet they face challenges with error accumulation. Consequently, the field has shifted toward holistic generation, which fine-tunes single-shot baselines (\textit{e.g.}, Wan~\cite{wan2025}) to model the entire sequence jointly~\cite{Guo2025LCT,xiao2025captain,Cai2025MOC,jia2025moga,multishotmaster,kara2025shotadapter,shotstream}.

\noindent\textit{\textbf{Camera Trajectory Datasets.}} MVImgNet~\cite{MVImgNet}, RealEstate-10K~\cite{RealEstate10k}, and DL3-DV-10K~\cite{DL3DV-10K} primarily capture basic paths in static environments, lacking cinematic narratives. While CCD~\cite{CCD} and E.T.~\cite{ET} introduce human-centric tracking, and GenDoP~\cite{gendop} focuses on cinematic moves, they remain confined to single-shot settings without global spatial coherence. \textit{In contrast, our \textbf{\textit{ShotVerse-Bench}} dataset uniquely provides multi-shot sequences calibrated into a unified coordinate system, serving as the first dataset for learning cross-shot spatial logic and cinematic consistency.}

\noindent\textit{\textbf{Camera Trajectory Generation.}} Early approaches relied on optimization~\cite{DBLP:journals/cga/Blinn88a,DBLP:journals/tog/LinoC15,DBLP:conf/mig/GalvaneCLR15,SplaTraj} or heuristic constraints~\cite{DBLP:journals/jfr/BonattiWHAGCKCS20,DBLP:conf/si3d/DruckerGZ92,DBLP:conf/cvpr/HuangLYK00C19}. Recent generative methods often depend on explicit 3D priors: CCD~\cite{CCD} and E.T.~\cite{ET} require character motion inputs, Director3D~\cite{Director3D} relies on object-centric data, and GenDoP~\cite{gendop} necessitates RGBD information. \textit{Unlike these methods that demand complex pre-construction, we utilize the strong semantic-spatial priors of VLMs to automate cinematic plotting.}

\noindent\textit{\textbf{Camera Control for Text-Driven Video Generation.}} Early approaches~\cite{he2024cameractrl,motionctrl,animatediff} encode camera extrinsics into pre-trained models. Subsequent works enhance geometric fidelity by using 3D priors~\cite{ac3d,vd3d,cami2v}. However, they lack the capability to model multi-shot storytelling. Recent research has expanded into synchronized multi-camera generation~\cite{baisyncammaster,recammaster,CVD,he2025cameractrl}, aiming for 3D-consistent scene modeling across different viewpoints. However, these methods ignore orchestrating cinematic cuts between distinct shots. The relevant work, ShotDirector~\cite{shotdirector}, pioneers shot transitions in single-shot video generation models by combining camera conditioning with editing-pattern-aware prompting. However, its trajectory patterns are largely constrained to fixed-point shooting or specific editing templates. \textit{In contrast, we propose a more cinematic dataset, systematically unify planning and controlling, and specifically target a holistic multi-shot video model.}
\vspace{-0.1cm}

\section{Methodology: ShotVerse}

We propose \textit{\textbf{ShotVerse}} (Fig.~\ref{fig:framework}) to decouple the multi-shot camera control task into planning and controlling phases, which bridges the gap between unreliable textual camera control and labor-intensive manual plotting. Guided by our data-centric perspective that aligned triplets of \textit{(Caption, Trajectory, Video)} form a joint distribution, this ``Plan-then-Control'' framework is presented as follows: First, Sec.~\ref{method:planner} details the Planner, which models the conditional probability \textit{P}(\textit{Trajectory} | \textit{Caption}) to synthesize cinematic, globally unified trajectories from hierarchical textual descriptions. Second, Sec.~\ref{method:controller} presents the Controller, modeling \textit{P}(\textit{Video} | \textit{Caption}, \textit{Trajectory}) to render these plotted trajectories into multi-shot videos using a holistic Diffusion Transformer~\cite{dit,holocine}.

\vspace{-0.2cm}
\subsection{Planner: Shot-Aware Cinematic Trajectory Plotting}
\label{method:planner}
\vspace{-0.1cm}
\noindent
\textbf{\textit{Motivation.}}
To achieve automated and cinematic trajectory generation, we must address critical limitations in existing paradigms. First, workflows~\cite{ET,CCD,Director3D} relying on pre-construction (\textit{e.g.}, 3D scene layout or character proxies) inherently lack the potential for scalability and automation. Second, while data-driven methods offer flexibility, previous works (\textit{e.g.}, GenDoP~\cite{gendop}) typically utilize shallow text encoders, which fail to capture the deep spatial reasoning required for complex multi-shot narratives. Large Vision-Language Models (\textit{e.g.}, Qwen3-VL~\cite{Qwen3-VL}) offer a promising solution to bridge this semantic-geometric gap due to their rich spatial priors~\cite{cai2025scaling}. However, a fundamental challenge remains: directly tasking a VLM to predict variable-length, long-horizon trajectory sequences is structurally inefficient and prone to model degeneration.

\noindent
\textbf{\textit{Overview.}} To reconcile these challenges, we propose a shot-aware cinematic trajectory Planner. As shown in Fig.~\ref{fig:framework}, we first construct a hierarchical prompt, and then adopt a VLM to encode these inputs to extract context-aware ``camera codes''. A trajectory decoder is adopted to expand these codes into variable-length trajectory tokens. Finally, a pose de-tokenizer de-tokenizes them into explicit camera poses. The Planner is trained end-to-end to maximize the log-likelihood of the ground-truth tokens.

\noindent
\textbf{\textit{Task Formulation and Trajectory (De-)Tokenization.}}
We formulate planning as generating a sequence of $K$ camera trajectories $\mathcal{S} = \{ \mathcal{S}^{(k)} \}_{k=1}^K$, where each shot $\mathcal{S}^{(k)} = \{ \mathbf{P}_t^{(k)} \}_{t=1}^{L_k}$ consists of $L_k$ poses defined in a unified global coordinate system. Each pose $\mathbf{P}_t^{(k)} \in SE(3)$ is represented using a 12D continuous vector $[\mathbf{t}_t, \mathbf{r}_t]$ where translation $\mathbf{t}_t \in \mathbb{R}^3$ and rotation $\mathbf{r}_t\in \mathbb{R}^9$. We implement a reversible tokenization pipeline where continuous camera parameters are normalized and discretized into integer bins. Conversely, the de-tokenizer maps generated tokens back to continuous values via bin centers, applying inverse scaling to recover the explicit trajectory in the unified global coordinate system.

\noindent
\textbf{\textit{Hierarchical Prompt Construction.}} To avoid semantic confusion between shots, we design a structured input sequence $\mathbf{I}_{in}$ that interleaves semantic context with learnable query placeholders as:
{\footnotesize
\begin{equation}
    \mathbf{I}_{in} = \text{Tok}(\mathcal{X}_{global}) \oplus \bigoplus_{k=1}^{K} \left( \text{Tok}(\mathcal{X}_{shot}^{(k)}) \oplus \underbrace{[\texttt{<TRAJ>}^{(k)}_1, \dots, \texttt{<TRAJ>}^{(k)}_{M}]}_{\text{Query Tokens for Shot } k} \right),
\end{equation}
}where $\oplus$ denotes concatenation, and $\text{Tok}(\cdot)$ denotes text tokenizer. The $\mathcal{X}_{text}$ is decomposed into a global prompt $\mathcal{X}_{global}$ and a sequence of per-shot prompts $\{\mathcal{X}_{shot}^{(1)}, \dots, \mathcal{X}_{shot}^{(K)}\}$. We append a fixed number of learnable trajectory query tokens, denoted as $\{\texttt{<TRAJ>}_m^{(k)}\}_{m=1}^{M}$, immediately after each shot's textual description. These tokens serve as ``slots'' for the VLM to fill with shot-specific camera plans.

\noindent
\textbf{\textit{Context-Aware Encoding.}} Since shallow text encoders~\cite{rombach2022high} lack spatial reasoning, we leverage the VLM backbone to perform a ``mental simulation'' of camera movement. Facilitated by the self-attention mechanism, the hidden states of the trajectory query tokens aggregate information from the global context, previous shots (causal dependency), and the specific instruction for the current shot. Specifically, we employ the VLM backbone $\Phi(\cdot)$ to encode the entire sequence $\mathbf{I}_{in}$, and extract the final-layer hidden states corresponding to the query tokens of each shot $k$ as camera codes, denoted as $\mathbf{H}_{plan}^{(k)} \in \mathbb{R}^{M \times D_{\text{vlm}}}$, where $D_{\text{vlm}}$ is the hidden dimension of the VLM.

\noindent
\textbf{\textit{Trajectory Decoder.}}
The VLM produces a fixed number ($M$) of camera codes for each shot, whereas the actual camera trajectory requires a variable-length sequence of pose tokens. To enable joint temporal modeling across shots, we concatenate all shot-level codes with learnable separator tokens $\texttt{<SEP>}$ as:
{\footnotesize
\begin{equation}
\mathbf{H}_{plan} =
\left[
\mathbf{H}_{plan}^{(1)}; \texttt{<SEP>}; \mathbf{H}_{plan}^{(2)}; \texttt{<SEP>}; \dots; \texttt{<SEP>};\mathbf{H}_{plan}^{(K)}
\right].
\end{equation}
}

\noindent
The concatenated sequence $\mathbf{H}_{plan}$ serves as a prefix to a lightweight auto-regressive Transformer~\cite{OPT} decoder, which models the probability $P(\mathbf{S}_{traj} | \mathbf{H}_{plan})$. Since the VLM hidden dimension may differ from the decoder embedding dimension, 
we project $\mathbf{H}_{\text{plan}}$ to the decoder embedding space via a linear layer 
to match the decoder embedding dimension $D$. At decoding step $j$, the input is formulated as:
{
\footnotesize
\begin{equation}
\mathbf{X}_j =
\mathrm{PosEmbed}
\left(
[\mathbf{H}_{plan}; \mathcal{V}[y_{0:j-1}]
\right)
\in
\mathbb{R}^{(\text{length}(\mathbf{H}_{plan}) + j) \times D},
\end{equation}
}
where $\mathcal{V}$ is a learnable codebook and $y_{0:j-1}$ are previously generated trajectory token IDs. The decoder applies stacked causal self-attention layers to predict the next token ${y}_j$ via a linear projection layer, enabling temporal modeling both within and across shots for cinematic continuity. This process ultimately generates a trajectory token sequence $\hat{\mathbf{S}}_{traj}
=
\{\hat{y}_j\}_{j=1}^{N+2}$, where $N$ is the variable trajectory sequence length, with two special tokens for beginning and ending.

\noindent
\textbf{\textit{Training Objective.}} We optimize the VLM parameters (via LoRA~\cite{hu2022lora}) and the Decoder using:
{
\footnotesize
\begin{equation}
\mathcal{L}_{plan}
=
\mathrm{CrossEntropy}(\mathbf{S}_{traj}, \hat{\mathbf{S}}_{traj})
+
\lambda \lVert \mathbf{H}_{plan} \rVert_2^2,
\end{equation}
}
where the second term applies L2 regularization to the latent code $\mathbf{H}_{plan}$ to prevent overfitting and ensure representation compactness.
\vspace{-0.6cm}

\subsection{Controller: Cinematic Camera Control for Text-Driven Multi-Shot Video Generation}
\label{method:controller}
\noindent
\textbf{\textit{Motivation.}} While the Planner provides precise geometric instructions, executing them poses a significant challenge: complex multi-shot cinematic trajectories (\textit{e.g.}, rapid cuts, variable-speed tracking) represent unseen patterns for pre-trained video models. Consequently, fine-tuning is essential to bridge this domain gap. However, naive fine-tuning often degrades visual quality or fails to capture the sharp temporal boundaries of cuts. To address these challenges, we propose a trajectory-conditioned Controller.

\noindent \textbf{\textit{Overview.}}
We employ the holistic multi-shot foundation model~\cite{holocine} for video synthesis. To advance its camera control, we employ a lightweight fine-tuning strategy (via LoRA) to adapt the model to new control signals and patterns. Our architecture adaptation is designed with two guiding principles: (i) a ``simple-yet-effective'' injection mechanism via a Camera Encoder that enforces geometric adherence without disrupting pre-trained priors; and (ii) a shot-aware structural bias via 4D Rotary Positional Embedding that explicitly informs the model of hierarchical boundaries. Finally, we optimize the controller using a Flow Matching objective.

\noindent
\textbf{\textit{Camera Encoder.}} We implement a direct feature injection mechanism~\cite{recammaster} to guide the camera trajectory understanding of the model. Specifically, for each frame $t$, the extrinsic matrix $\mathbf{E}_t \in \mathbb{R}^{3 \times 4}$ is first flattened and projected to match the video token channels $d$ via a learnable Camera Encoder $\mathcal{E}_c$ (instantiated as a Fully Connected Layer), $\mathbf{c}_{cam} = \mathcal{E}_c(\text{Flatten}(\mathbf{E}_t)) \in \mathbb{R}^{d}$. To achieve fine-grained control, this encoder is inserted into each transformer block. We directly add the camera embedding to the intermediate visual features. Specifically, let $\mathbf{F}_{norm}$ denote the features after layer normalization and adaptive modulation. The input features, $\mathbf{F}_{attnin}$, for the self-attention layer are computed as: $\mathbf{F}_{attnin} = \mathbf{F}_{norm} + \mathbf{c}_{cam}$. By injecting the trajectory signal $\mathbf{c}_{cam}$ right before the self-attention layer, we explicitly condition the temporal modeling process on the target camera pose, ensuring the generated motion dynamics align with the camera condition. While direct injection proves effective in video-to-video settings~\cite{recammaster} with synthetic data, applying it to text-driven generation risks texture distribution drift. We resolve this via our data-centric foundation: by training on the real-world \textbf{\textit{ShotVerse-Bench}} dataset, we ensure the model aligns camera control with natural cinematic textures.

\begin{table*}[t]
\centering
\caption{\textbf{Comparisons of Camera Trajectory Datasets.} \textbf{\textit{ShotVerse-Bench}} is the first large-scale dataset that provides multi-shot cinematic camera trajectories together with rich, multi-level caption annotations.}
\vspace{-0.4cm}
\setlength{\tabcolsep}{10pt}
\resizebox{\linewidth}{!}{
\begin{tabular}{lccccccccc}
\toprule
\multirow{2}{*}{Dataset} & \multicolumn{4}{c}{Caption Annotation} & \multirow{2}{*}{Traj Type} & \multirow{2}{*}{Domain} & \multicolumn{2}{c}{Statistics} \\
\cmidrule(lr){2-5}\cmidrule(lr){8-9}
 & Traj & Scene & Intent & \#Vocab &  &  & \#Sample & \#Frame  \\
\midrule
MVImgNet~\cite{MVImgNet} & $\times$ & $\times$ & $\times$ & -- & Object/Scene-Centric & Captured & 22K & 6.5M \\

RealEstate10k~\cite{RealEstate10k} & $\times$ & $\times$ & $\times$ & -- & Object/Scene-Centric & Youtube & 79K & 11M \\

DL3DV-10K~\cite{DL3DV-10K} & $\times$ & $\times$ & $\times$ & -- & Object/Scene-Centric & Captured & 10K & \textbf{51M} \\

CCD~\cite{CCD} & \textbf{$\checkmark$} & $\times$ & $\times$ & 48 & Tracking & Synthetic & 25K & 4.5M \\

E.T.~\cite{ET} & \textbf{$\checkmark$} & $\times$ & $\times$ & 1790 & Tracking & Film & \textbf{115K} & 11M \\

GenDoP~\cite{gendop} & \textbf{$\checkmark$} & \textbf{$\checkmark$} & \textbf{$\checkmark$} & 8698 & Free-Moving & Film & 29K& 11M \\

ShotVerse-Bench (Ours) & \textbf{$\checkmark$}  & \textbf{$\checkmark$}  & \textbf{$\checkmark$}  & \textbf{19819}  & Free-Moving\&\textbf{Multi-Shot} & Film/\textbf{TV}/\textbf{Documentary} & 20.5K & 12M \\
\bottomrule
\end{tabular}}
\label{tab:dataset}
\end{table*}

\noindent
\textbf{\textit{4D Rotary Positional Embedding.}} Standard video generation models typically utilize 3D positional embeddings (\textit{i.e.}, frame, height, width). However, multi-shot videos possess a hierarchical temporal structure (\textit{i.e.}, video $\to$ shot $\to$ frame). To explicitly inform the model about shot boundaries and enforce intra-shot consistency, we propose a 4D Rotary Positional Embedding (4D RoPE) strategy. This process operates in three streamlined steps: (i) \textit{Dimension Allocation.} We partition the attention head dimension into four subspaces: $F_{shot},F_{frame}, F_h,F_w,$. We allocate a larger proportion to spatial dimensions ($F_h,F_w$) to preserve visual fidelity, while reserving sufficient capacity in $F_{shot}$ and $F_{frame}$ for hierarchical temporal modeling. (ii) \textit{Frequency Pre-Computation.} Independent rotary frequency banks are pre-calculated for each dimension following the standard RoPE formulation. This creates orthogonal positional bases for height, width, shot index, and frame index. (iii) \textit{Dynamic Assembly.} During the forward pass, we dynamically map each video frame to its corresponding shot index $s$ and  global frame index $t$. The final embedding is assembled by concatenating the frequency components from all subspaces. Crucially, this mechanism ensures that all frames within the same shot share a unified shot embedding, explicitly enforcing intra-shot consistency while time handles fine-grained temporal dynamics.

\noindent
\textbf{\textit{Training Objective.}} Following the HoloCine training protocol, we employ the Flow Matching~\cite{Lipman2022FlowMF} objective. Given a clean video latent $\mathbf{v}_0$ and a Gaussian noise sample $\mathbf{v}_1 \sim \mathcal{N}(\mathbf{0}, \mathbf{I})$, we interpolate between them to obtain $\mathbf{v}_\sigma = (1 - \sigma)\mathbf{v}_0 + \sigma\mathbf{v}_1$ at noise level $\sigma \in [0,1]$. The model $v_\theta$ is trained to predict the velocity field $\mathbf{u} = \mathbf{v}_1 - \mathbf{v}_0$. The training objective is formulated as follows:

{\footnotesize
\begin{equation}
    \mathcal{L}_{control} = \mathbb{E}_{\sigma, \mathbf{v}_0, \mathbf{v}_1} \left[ \| v_{\theta}(\mathbf{v}_\sigma, \sigma, \mathbf{c}_{text}, \mathbf{c}_{cam}) - (\mathbf{v}_1 - \mathbf{v}_0) \|_2^2 \right],
\end{equation}
}
where $\mathbf{c}_{text}$ represents the textual embeddings extracted by the text encoder, and $\mathbf{c}_{cam}$  denotes the explicit camera condition embeddings.

\section{Dataset and Benchmark: ShotVerse-Bench}
\subsection{Dataset Curation}
\label{method:datasets}

\noindent
\textbf{\textit{Dataset Overview.}} A core challenge in camera-controlled multi-shot cinematic video generation is the lack of data that aligns semantic descriptions with globally unified camera trajectory annotations. To address this, we construct a dataset, \textbf{\textit{ShotVerse-Bench}}, featuring hierarchical captions and unified multi-shot trajectories (Tab.~\ref{tab:dataset}). We collect 20,500 clips from high-production cinema, ensuring that the content adheres to professional cinematic standards and cinematography principles, which covers a broad and balanced taxonomy of camera control.

\noindent
\textbf{\textit{Quantitative Dataset Diversity.}} Beyond its 27+ fine-grained trajectory categories, ShotVerse-Bench covers diverse camera motions and compositions, including 1,639 zoom/scale, 463 translation, 158 rotation/arc, 1,295 complex/reveal, 4,874 close-up, 3,543 medium-range, and 512 wide/long-range samples. It further contains 1,720 explicit cuts or transitions and spans TV series (10,864 clips), movies (3,962), animation (3,441), and documentaries (2,233). The clips have a mean duration of 5.50 seconds and a maximum duration of 19.98 seconds, demonstrating diversity across camera motion, framing scale, temporal extent, and content domain.

\noindent
\textbf{\textit{Multi-Shot Camera Calibration.}} To align disjoint single-shot trajectories into a unified global coordinate system, we propose a four-step calibration pipeline. (i) \textit{Dynamic Foreground Removal.} To address the dynamic objects, we employ SAM~\cite{sam2} to mask foregrounds, retaining static background regions for robust pose estimation. (ii) \textit{Single-Shot Local Reconstruction.} We then independently reconstruct each shot $s$ using PI3~\cite{pi3} on the static background, producing a locally consistent trajectory within a shot-specific local frame. (iii) \textit{Joint Keyframe Global Reconstruction.} We sample keyframes across disjoint shots and reconstruct them jointly via PI3. This yields a unified static scene and global poses, naturally defining a global coordinate system for the entire multi-shot sequence. (iv) \textit{Anchor-Based Trajectory Alignment.} To unify trajectories, we identify an anchor frame for each shot present in both local and global reconstructions, yielding dual pose references. We estimate a similarity transformation to align the local frame to the global system, resolving scale ambiguity by comparing the relative displacements of the shot's start and end keyframes.

\noindent
\textbf{\textit{Training and Test.}} To construct multi-shot training data from the single-shot clips, we assemble multi-shot sequences of 249 frames. We select 2,750 representative single-shot clips and group them into 1,100 multi-shot scenes, each containing 2, 3, or 4 shots following a 6:3:1 ratio. After removing embedded subtitles and standardizing resolution to $832\times480$, we allocate 1,000 scenes for training and 100 for testing with no scene overlap.

\subsection{Evaluation Benchmark: A Three-Track Protocol}

\label{subsec:benchmark}
Our target task is \emph{text-driven cinematic multi-shot video creation with explicit, globally unified camera control}, which requires models to jointly solve trajectory planning and faithful execution.
To our knowledge, this task has not been systematically evaluated by prior benchmarks; therefore, we introduce a three-track protocol for comprehensive evaluation at both component and system levels, by separately measuring (A) text-to-trajectory planning, (B) trajectory-to-video execution fidelity, and (C) end-to-end text-to-video generation quality.

\noindent \textit{\textbf{Track A: Text-to-Trajectory.}} 
This track evaluates the Planner's ability to translate narrative intent into explicit camera trajectories. (i) \textit{Input/Output}: Hierarchical prompts $\to$ Globally aligned camera sequences. (ii) \textit{Alignment Metrics}: We adopt \textit{F1-Score} for discrete motion tag alignment and \textit{CLaTr-CLIP}~\cite{gendop} for soft semantic alignment.

\noindent \textit{\textbf{Track B: Trajectory-to-Video.}} 
This track assesses the Controller's execution fidelity given ground-truth trajectories. (i) \textit{Input/Output}: Trajectories + Prompts $\to$ Multi-shot video. (ii) \textit{Control Accuracy}: We use PI3~\cite{pi3} to extract poses from generated videos and compute \textit{Transition Error} and \textit{Rotation Error} against ground truth. (iii) \textit{Coordinate Alignment}: We further introduce the \textit{Coordinate Alignment Score} (CAS), which selects cross-shot frame pairs with the highest field-of-view overlap and measures their visual consistency via DINOv2~\cite{oquab2023dinov2} similarity---if the coordinate system is well unified, frames with higher geometric overlap are expected to exhibit higher visual similarity. 

\noindent \textit{\textbf{Track C: Text-to-Video.}} 
This track measures the end-to-end performance of integrated planning and execution. (i) \textit{Input/Output}: Hierarchical prompts $\to$ Multi-shot video. (ii) \textit{Semantic Consistency}: We assess \textit{Global and Shot-level Consistency} via pairwise ViCLIP~\cite{wang2023internvid} embeddings. (iii) \textit{Visual Quality}: We report \textit{Aesthetic Quality} (LAION predictor~\cite{laion_aesthetic}), \textit{Shot Transition Accuracy} for temporal cut precision, and \textit{FVD} for temporal coherence. (iv) \textit{Cinematic Planning Quality}: We conduct VLM-based (Gemini 3 Pro\cite{gemini}) and user studies across four dimensions: \textit{Motion Type Appropriateness}, \textit{Motion Duration Appropriateness}, \textit{Subject Emphasis \& Saliency}, and \textit{Cinematic Pacing}.

\subsection{Baseline Selection}
We compare \textit{ShotVerse} against: (i) \textit{Trajectory Planners (Track A):} We compare with representative camera trajectory generation methods, including 
CCD~\cite{CCD}, 
E.T.~\cite{ET}, 
Director3D~\cite{Director3D}, 
and GenDoP~\cite{gendop}. 
Among them, GenDoP serves as the strongest autoregressive trajectory generation baseline. 
For multi-shot evaluation, these methods are applied under the same hierarchical prompt setting for fair comparison. (ii) \textit{Camera-Controlled Baselines (Track B):} We evaluate state-of-the-art single-shot control models, including CameraCtrl~\cite{he2024cameractrl}, MotionCtrl~\cite{motionctrl}, and ReCamMaster~\cite{recammaster}. To adapt them for multi-shot evaluation, we apply these models shot-by-shot and concatenate the results using our proposed calibration pipeline for alignment. (iii) \textit{Multi-Shot Video Models (Track C):} (iii-a) \textit{Open-Source Models}: We compare against HoloCine~\cite{holocine} and MultiShotMaster~\cite{multishotmaster}. (iii-b) \textit{Closed-Source Models}: We include leading closed-source models, which are Sora2~\cite{sora2}, VEO3~\cite{veo}, Kling3.0~\cite{kling}, and Seedance2.0~\cite{gao2025seedance}. As these models rely on implicit textual control, we provide them with our hierarchical prompts to evaluate their zero-shot cinematic understanding.

\section{Experiments}

\subsection{Implementation Details}

The Planner integrates a Qwen3-VL-2B backbone with an OPT-based decoder (12 layers), utilizing LoRA ($r=32$) and discrete tokenization ($B=256$) for trajectory synthesis. Inference employs Nucleus sampling ($\tau=0.9, p=0.95$). The Controller adapts HoloCine via rank-128 LoRA and a fully-connected camera encoder. Following the Wan 2.2 protocol~\cite{wan2025}, we employ a critical two-stage training strategy: the camera encoder is optimized only during the high-noise stage ($0.875\le \sigma \le 1$) to anchor coarse motion, while the low-noise stage refines details via LoRA only. Optimization uses AdamW (learning rate $=10^{-4}$) on 96 NVIDIA H20 GPUs with FSDP~\cite{zhao2023pytorchfsdpexperiencesscaling}.

\subsection{Benchmark Results}

Results follow the three-track protocol (Sec.\ref{subsec:benchmark}), focusing on alignment, control, and cinematic quality. Recognizing that numerical trajectory metrics often fail to reflect the actual cinematographic experience, we emphasize the evaluation of rendered video outputs. We employ a dual-pronged assessment strategy combining VLM-based scoring (Gemini 3 Pro\cite{gemini}) with diverse human user studies.

\noindent
\textbf{\textit{Track A: Text-to-Trajectory.}} Tab.~\ref{tab:align} compares our Planner against CCD~\cite{CCD}, E.T.~\cite{ET}, Director3D~\cite{Director3D}, and GenDoP~\cite{gendop}. CCD, E.T., and Director3D underperform significantly, indicating that existing trajectory generation methods struggle with complex, especially multi-shot, narrative prompts. GenDoP, the strongest baseline, achieves competitive results on its native DataDoP benchmark but suffers a notable domain gap when re-trained on \textit{ShotVerse-Bench}. Our VLM-driven Planner achieves the best results on both benchmarks, demonstrating stronger cross-domain generalization from the vision-language backbone.

\begin{table}[t]
    \centering
    \caption{\textbf{Track A: Quantitative Results of Text-Trajectory Alignment.}}
          \label{tab:align}
          \vspace{-0.4cm}
          \resizebox{\linewidth}{!}{
          \begin{tabular}{lccccc}
            \toprule
            \multirow{2}{*}{\textbf{Method}} & \multirow{2}{*}{\textbf{Dataset}}& \multicolumn{2}{c}{DataDoP~\cite{gendop}} & \multicolumn{2}{c}{ShotVerse-Bench} \\
            \cmidrule(lr){3-4} \cmidrule(lr){5-6}
            & & F1-Score $\uparrow$ & CLaTr-CLIP $\uparrow$ & F1-Score $\uparrow$ & CLaTr-CLIP $\uparrow$ \\
            \midrule
            CCD~\cite{CCD}&Pre-Trained&0.315&4.247&0.323&7.032\\
            E.T.~\cite{ET}&Pre-Trained&0.319&0.000&0.289&2.462\\
            Director3D~\cite{Director3D}&Pre-Trained&0.126&0.000&0.162&1.237\\
            GenDoP~\cite{gendop} &Pre-Trained& 0.399 & 32.408 & 0.326 & 23.089 \\
            GenDoP~\cite{gendop} &ShotVerse-Bench& 0.268 & 24.132 & 0.343 & 33.875 \\
            
            ShotVerse (Ours) &ShotVerse-Bench& \textbf{0.418} & \textbf{34.907} & \textbf{0.422} & \textbf{35.016} \\
            \bottomrule
          \end{tabular}}
        \vspace{-0.15cm}
\end{table}

\begin{table}[t]
    \centering
    \caption{\textbf{Track B: Quantitative Evaluation of Camera Control.} All methods receive ground-truth trajectories.}
          \label{tab:controller}
          \vspace{-0.4cm}
          \resizebox{\linewidth}{!}{\begin{tabular}{lccc}
            \toprule
            Method &
            Trans. Error $\downarrow$ &
            Rotation Error $\downarrow$ &
            CAS $\uparrow$ \\
            \midrule
            MotionCtrl~\cite{motionctrl} & 0.0900 & 2.56 & 0.329\\
            CameraCtrl~\cite{he2024cameractrl} & 0.0571 & 1.28 & 0.343\\
            ReCamMaster~\cite{recammaster} & 0.0589 & 1.12 & 0.408\\
            
            ShotVerse (Ours) & \textbf{0.0163} & \textbf{0.73} & \textbf{0.500}\\
            \bottomrule
          \end{tabular}}
          \vspace{-0.15cm}
\end{table}

\begin{table}[t]
    \centering
    \caption{\textbf{Track C: Quantitative Evaluation of Multi-Shot Quality.} Without shot-splitting, shot metrics cannot be calculated for some baselines.}
          \label{tab:trackc}
          \vspace{-0.4cm}
          \resizebox{\linewidth}{!}{\begin{tabular}{lccccc}
            \toprule
            Method &
           \makecell{Sem. Consist. \\ (Global) $\uparrow$} &  \makecell{Sem. Consist. \\(Shot) $\uparrow$}&
            \makecell{Aesthetic\\ Quality $\uparrow$} &
            \makecell{Shot Trans.\\ Accuracy $\uparrow$} &
            \makecell{FVD $\downarrow$}\\
            \midrule
            HoloCine~\cite{holocine} & 0.297 & 0.254 & 4.981 & 0.645 & 407.54\\
            MultiShotMaster~\cite{multishotmaster} & 0.279 & 0.247 & 5.210 & 0.927 & 440.78\\
            Sora2~\cite{sora2} & 0.297 & - & 5.344 & - & 372.13\\
            VEO3~\cite{veo} & 0.282 & - & 5.441 & - & 941.50\\
            Kling3.0~\cite{kling} & 0.288 & - & 5.167 & - & 719.44\\
            Seedance2.0~\cite{gao2025seedance} & 0.285 & - & 5.381 & - & 605.17\\
            
            ShotVerse (Ours) & \textbf{0.299} & \textbf{0.255} & \textbf{5.465} & \textbf{0.933} & \textbf{281.71}\\
            \bottomrule
          \end{tabular}}
          \vspace{-0.15cm}
\end{table}

\begin{table}[t]
    \centering
    \caption{\textbf{Track C: Quantitative Evaluation of Cinematic Quality.}}
           \vspace{-0.4cm}
          \label{tab:planner}
          \resizebox{\linewidth}{!}{\begin{tabular}{clcccc}
            \toprule
            &Method &
            \makecell{Motion Type \\ Appropriateness $\uparrow$} &
            \makecell{Motion Duration \\ Appropriateness $\uparrow$} &
            \makecell{Subject Emphasis \\ \& Saliency $\uparrow$} &
            \makecell{Cinematic \\ Pacing $\uparrow$} \\
            \midrule
            &HoloCine~\cite{holocine} & 4.324 & 4.281 & 3.997 & 3.208 \\
            &VEO3~\cite{veo} &4.402 & 4.189 & 4.252 & 3.288 \\
            VLM-Based&Sora2~\cite{sora2} & 4.371 & 4.258 & 3.892 & 3.236 \\
            &Kling3.0~\cite{kling} & 4.302 & 4.153 & 3.872 & 3.108 \\
            &Seedance2.0~\cite{gao2025seedance} & 4.402 & 4.279 & 4.328 & 3.279 \\
            
            &ShotVerse (Ours) & \textbf{4.447} & \textbf{4.304} & \textbf{4.426} & \textbf{3.384} \\
            \midrule
            &HoloCine~\cite{holocine} & 2.555 & 2.615 & 2.585 & 2.545 \\
            &VEO3~\cite{veo} & 3.564 & 3.892 & 3.649 & 3.561 \\
            User Study&Sora2~\cite{sora2} & 3.665 & 3.645 & 3.865 & 3.625 \\
            &Kling3.0~\cite{kling} & 3.702 & 3.572 & 3.598 & 3.539 \\
            &Seedance2.0~\cite{gao2025seedance} & 3.987 & 3.820 & 3.703 & 3.974 \\
            
            &ShotVerse (Ours) & \textbf{4.105} & \textbf{4.060} & \textbf{4.240} & \textbf{4.055} \\
            \bottomrule
          \end{tabular}}
          \vspace{-0.1cm}
\end{table}

\noindent
\textbf{\textit{Track B: Trajectory-to-Video.}} Tab.~\ref{tab:controller} compares camera control fidelity given ground-truth trajectories. Single-shot baselines MotionCtrl~\cite{motionctrl} and CameraCtrl~\cite{he2024cameractrl} exhibit high trajectory errors due to the lack of cross-shot coordination. ReCamMaster~\cite{recammaster} reduces rotation error but still suffers from coordinate misalignment across shots. Our Controller achieves the lowest errors on both translation and rotation, and the highest CAS, suggesting stronger cross-shot consistency under our CAS proxy.

\begin{figure*}[tb]
    \centering
    \includegraphics[width=\linewidth]{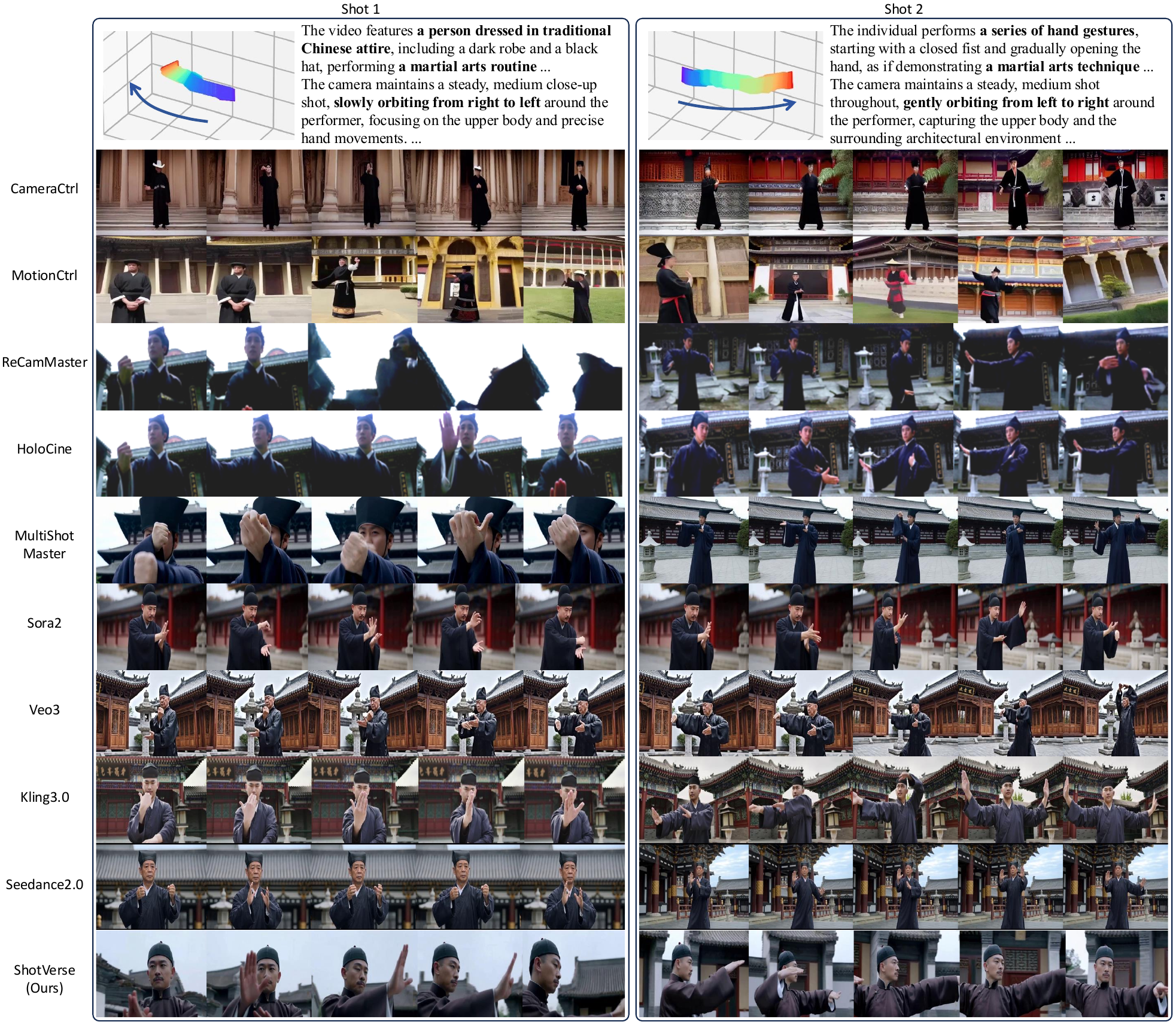}
    \vspace{-0.6cm}
    \caption{\textbf{Comparisons with the State-of-the-Art Baseline Methods.} Early camera-controlled text-driven generation models (\textit{e.g.}, CameraCtrl, MotionCtrl) struggle to handle complex cinematic camera trajectories. ReCamMaster executes the trajectory but drifts away from the subject in Shot 1. HoloCine, MultiShotMaster, Sora2, VEO3, and Kling3.0, and Seedance2.0~ fail to execute the complex ``orbit'' command, remaining nearly static. These failures demonstrate that for text-driven models, scaling up caption density is insufficient to achieve precise control without explicit geometric guidance.}
    \label{fig:compare}
    \vspace{-0.1cm}
\end{figure*}

\noindent
\textbf{\textit{Track C: Text-to-Video.}} Tab.~\ref{tab:trackc}, Tab.~\ref{tab:planner}, and Fig.~\ref{fig:compare} report end-to-end results. Our method achieves the lowest FVD and highest Aesthetic Quality (5.465), outperforming both open-source and commercial baselines. Notably, commercial models achieve competitive aesthetics but suffer from significantly higher FVD, indicating poor temporal fidelity without explicit trajectory guidance. For Shot Transition Accuracy, HoloCine (0.645) reflects the limitation of standard 3D positional encoding, MultiShotMaster (0.927) benefits from improved positional encoding, and our 4D RoPE further raises the score to 0.933 by explicitly modeling shot indices. For cinematic quality (Tab.~\ref{tab:planner}), both VLM-based and user study evaluations confirm our method leads across all four dimensions, with particularly strong gains in Subject Emphasis \& Saliency and Cinematic Pacing.

\vspace{-0.2cm}
\subsection{Ablation Study}

\noindent
\textit{\textbf{Planner.}} To analyze the contribution of the trajectory Planner, we conduct three structural ablations (Tab.~\ref{tab:ablation_planner}). (i) \textit{VLM encoder provides strong spatial priors}:  Replacing the VLM backbone with the original shallow encoder adopted in GenDoP (\textit{i.e.}, \textit{w/o VLM encoder}) leads to consistent degradation in both F1-Score and CLaTr-CLIP (Tab.~\ref{tab:ablation_planner}). VLM semantic-spatial priors are vital for multi-shot camera semantics; omitting them compromises the Planner's narrative alignment and degrades trajectory-text consistency. (ii) \textit{Dedicated trajectory decoder is necessary}: Directly adopting the VLM’s native decoder to autoregressively generate trajectory tokens significantly degrades performance. Our shot-aware Transformer decoder avoids the structural inefficiency of naive language modeling, ensuring stable long-horizon prediction through explicit temporal modeling. (iii) \textit{Query tokens enable shot-aware planning}: Removing the learnable trajectory query tokens (\textit{i.e.}, \textit{w/o Query Tokens}) and directly feeding the VLM encoder outputs to the decoder reduces F1-Score and CLaTr-CLIP. Query tokens act as structured ``planning slots'' to disentangle per-shot reasoning; without them, shared representations entangle semantics with geometry, compromising shot-specific control and alignment.

\begin{figure*}[t]
    \centering
    \includegraphics[width=0.95\linewidth]{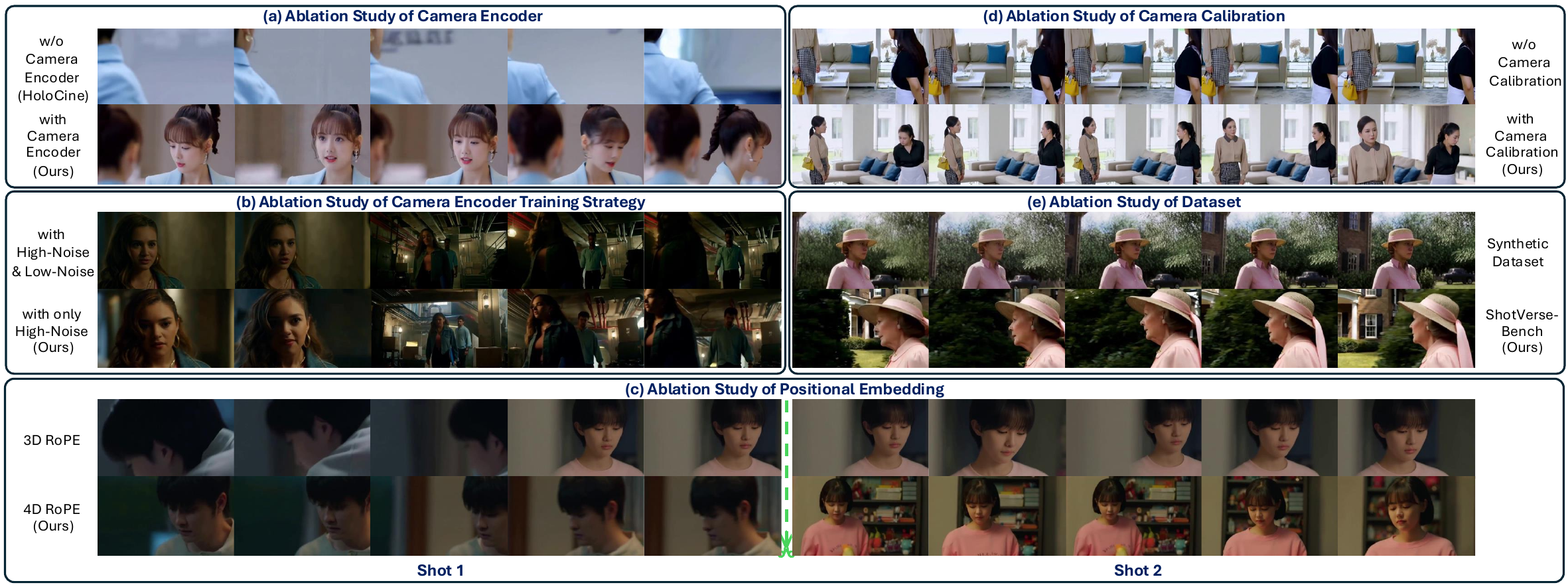}
    \vspace{-0.2cm}
    \caption{\textbf{Qualitative Ablation Study.}}
    \label{fig:ablation}
    \vspace{-0.1cm}
\end{figure*}

\noindent
\textit{\textbf{Controller.}} ShotVerse's gains come from (i) Controller-specific designs and (ii) curated data, instead of the video foundation model. (i-a) \textit{Camera encoder is necessary for controllability}: Fig.~\ref{fig:ablation} (a) shows that without the camera encoder (\textit{i.e.}, HoloCine), the model follows the intended motion pattern less reliably, whereas adding the encoder yields clearer, more stable camera behavior. (i-b) \textit{High-noise-only injection is sufficient}: Fig.~\ref{fig:ablation} (b) and Tab.~\ref{tab:ablation_controller} indicate that adding an additional low-noise encoder slightly trades off perceptual quality, implying early (high-noise) pose injection already establishes the global motion scaffold. (i-c) \textit{4D RoPE captures shot hierarchy}: Replacing 4D RoPE with 3D RoPE significantly degrades Shot Transition Accuracy from 0.933 to 0.429 (Tab.~\ref{tab:ablation_controller}), demonstrating that the explicit shot axis is critical for respecting shot boundaries. The ablation also changes the behavior around shot boundaries (Fig.~\ref{fig:ablation} (c)) and shifts the consistency/semantics trade-off, supporting that 4D RoPE helps preserve intra-shot coherence while respecting multi-shot structure. (ii-d) \textit{Unified camera calibration is necessary}: Removing global calibration reduces inter-shot consistency and aesthetics (Fig.~\ref{fig:ablation} (d)), supporting that unified coordinates are important for geometrically consistent pose conditioning across cuts. (ii-e) \textit{Synthetic supervision hurts film-like rendering}: Aesthetics drops noticeably and semantics slightly weakens (Tab.~\ref{tab:ablation_controller}), suggesting real cinematic triplets provide crucial cues beyond what synthetic triplets capture (Fig.~\ref{fig:ablation} (e)).

\begin{table}[t]
  \centering
  \scriptsize
  \caption{\textbf{Quantitative Evaluation of Ablation Study (Planner).}}
  \vspace{-0.4cm}
  \label{tab:ablation_planner}
  \resizebox{\linewidth}{!}{
  \begin{tabular}{lcccc}
    \toprule
    & w/o VLM encoder & w/ VLM decoder & w/o Query Tokens & Ours \\
    \midrule
    F1-Score $\uparrow$ & 0.343 & 0.248 & 0.251 & \textbf{0.422} \\
    CLaTr-CLIP $\uparrow$ & 33.875 & 15.078 & 18.796 & \textbf{35.016} \\
    \bottomrule
  \end{tabular}}
  \vspace{-0.2cm}
\end{table}

\begin{table}[t]
  \centering
  \scriptsize
  \caption{\textbf{Quantitative Evaluation of Ablation Study (Controller).}}
  \vspace{-0.4cm}
  \label{tab:ablation_controller}
  \resizebox{\linewidth}{!}{\begin{tabular}{lcccccc}
    \toprule
    Method &
    \makecell{Trans.\\ Error $\downarrow$ }&
    \makecell{Rotation \\Error $\downarrow$ }&
    \makecell{Shot Trans.\\ Acc. $\uparrow$}  &
  \makecell{Sem. Consist. \\ (Global) $\uparrow$} &  \makecell{Sem. Consist.\\ (Shot) $\uparrow$}&
    \makecell{Aesthetic\\ Quality $\uparrow$}\\
    \midrule
    w/o Cam. Enc. (HoloCine) & 0.0609 & 1.27 & 0.645 & 0.297 & 0.254 & 4.981\\
    w/ Low\&High Noise Enc. & 0.0189 & 0.74 & 0.930 & 0.296 & 0.250 & 5.321\\
    w/ 3D RoPE & 0.0323 & 1.04 & 0.429 & 0.290 & 0.251 & 5.413\\
    w/ Synthetic Data & 0.0509 & 1.35 & 0.705 & 0.292 & 0.253 & 4.833\\
    w/o Camera Calibration & 0.0165 & 0.79 & 0.931 & 0.296 & 0.251 & 5.136\\
    ShotVerse (Ours) & \textbf{0.0163} & \textbf{0.73} & \textbf{0.933} & \textbf{0.299} & \textbf{0.255} & \textbf{5.465}\\
    \bottomrule
  \end{tabular}}
  \vspace{-0.2cm}
\end{table}

\subsection{Discussion on Generalizability}

Beyond establishing state-of-the-art performance, our comprehensive error analysis yields three critical insights.

\begin{figure*}[t]
    \centering
    \includegraphics[width=0.85\linewidth]{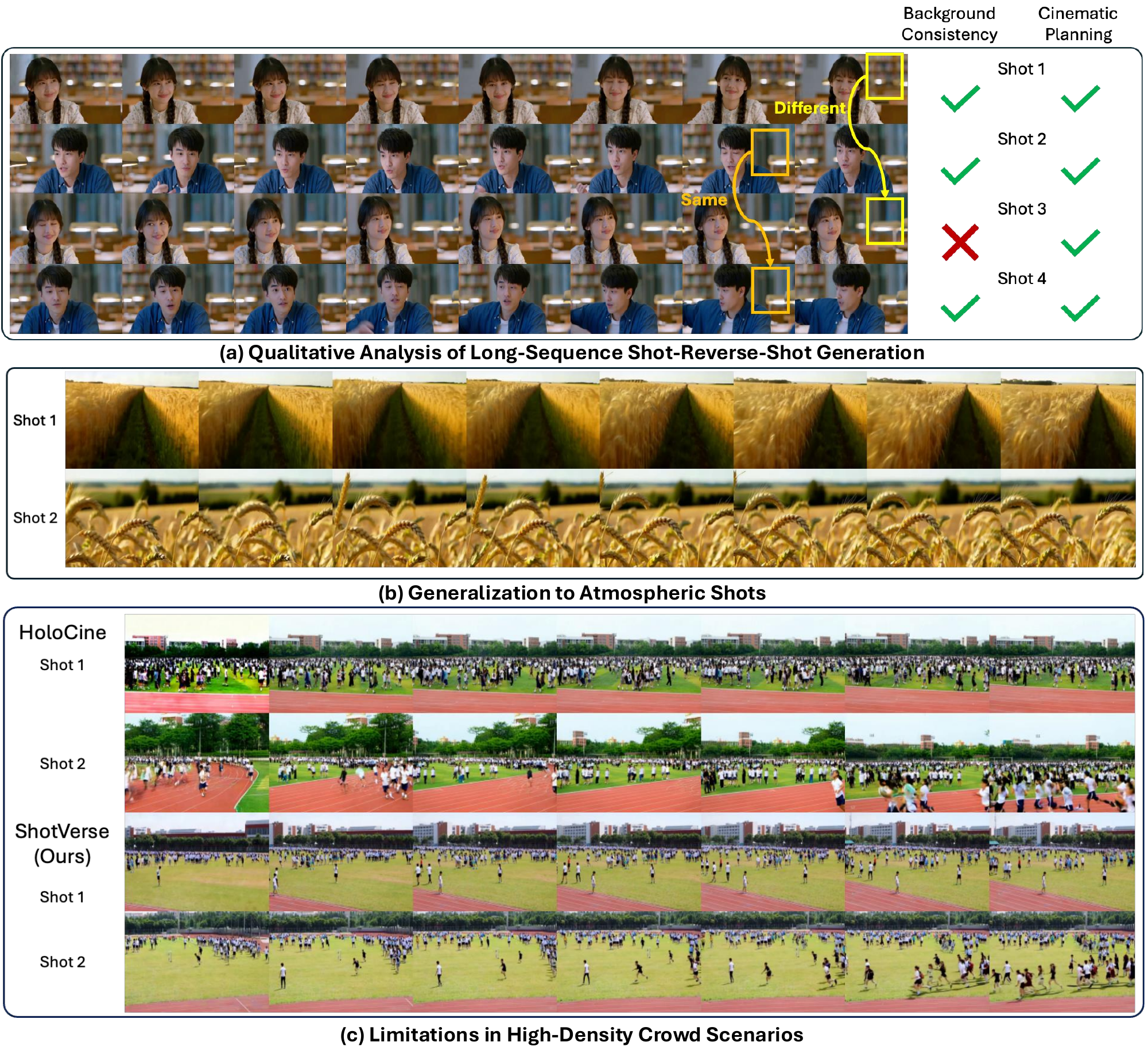}
    \vspace{-0.2cm}
    \caption{\textbf{Discussion on Generalizability}.}
    \label{fig:discussion}
    \vspace{-0.1cm}
\end{figure*}

\noindent
\textbf{\textit{Semantic-Geometric Synergy.}} We uncover a vital synergy in shot-reverse-shot scenarios, as shown in Fig.~\ref{fig:discussion} (a), where textual priors effectively compensate for calibration noise. However, minor drifts in long-context recurring views persist, signaling that achieving pixel-perfect scene persistence remains an open challenge.

\noindent
\textbf{\textit{Asymmetric Generalization.}} Our test split contains 100 unseen multi-shot scenes with no scene overlap with the training set. ShotVerse also transfers effectively to underrepresented atmospheric content (Fig.~\ref{fig:discussion} (b)), including aerial landscapes and empty environments. Nevertheless, borderline cases involving long shot-reverse-shot sequences or high-density crowds (Fig.~\ref{fig:discussion} (c)) reveal limitations in long-context persistence and dense multi-object generation: the Planner may occasionally produce off-target trajectories, while the Controller may exhibit background drift or subject flickering when camera motion interacts with many independently moving subjects and complex occlusions.

\noindent
\textbf{\textit{Autoregressive and Cross-Scene Extensions}.} Our method is currently limited to single-scene, multi-shot video, where the holistic “God-view” paradigm enables strong camera planning but constrains duration and flexibility. It is, however, compatible with autoregressive generation: intra-scene, trajectory conditions serve as global spatial anchors to reduce drift, while inter-scene, a VLM planner can generate consistent semantic tokens and a diffusion controller enforce pixel-level alignment. Extending this to longer, multi-scene generation remains future work.

\section{Conclusions}


In this work, we have pioneered a data-centric shift in the landscape of cinematic, multi-shot video generation. By introducing \textbf{\textit{ShotVerse}}, we bridge the long-standing gap between high-level narrative intent and low-level geometric precision, moving beyond simple video synthesis toward professional cinematographic orchestration. Our ``Plan-then-Control'' framework demonstrates that the complex spatial logic required for multi-shot storytelling can be effectively decoupled into a VLM-driven cognitive plotting phase and a geometry-aware rendering phase. Central to this breakthrough is the curation of \textbf{\textit{ShotVerse-Bench}}. By establishing a novel calibration pipeline that unifies disjoint shot trajectories into a global coordinate system, we provide the community with the first high-fidelity dataset capable of teaching AI the ``grammar of film''. Our extensive evaluation, conducted via a rigorous three-track protocol, confirms that \textbf{\textit{ShotVerse}} not only achieves state-of-the-art technical accuracy but also manifests a profound implicit understanding of cinematic pacing and visual salience.


\begin{acks}

This work was supported by grants from the NSFC/RGC Collaborative Research Scheme Project No. CRS\_HKUST605/25 and CCF-Tencent Rhino-Bird Young Faculty Open Research Fund. 

\end{acks}

\bibliographystyle{ACM-Reference-Format}
\bibliography{sample-base}

\clearpage 

\end{document}